\documentclass[conference]{IEEEtran}
\IEEEoverridecommandlockouts
\usepackage{cite}
\usepackage{amsmath,amssymb,amsfonts}
\usepackage{algorithmic}
\usepackage{graphicx}
\usepackage{textcomp}
\usepackage{xcolor}
\def\BibTeX{{\rm B\kern-.05em{\sc i\kern-.025em b}\kern-.08em
    T\kern-.1667em\lower.7ex\hbox{E}\kern-.125emX}}
\begin{document}

\title{Toward Transparent AI: A Survey on Interpreting\\the Inner Structures of Deep Neural Networks
}

\author{
\IEEEauthorblockN{Tilman Räuker$^*$}
\IEEEauthorblockA{
\textit{traeuker@gmail.com}
}
\and
\IEEEauthorblockN{Anson Ho$^*$}
\IEEEauthorblockA{\textit{Epoch} \\
\textit{anson@epochai.org}}

\and
\IEEEauthorblockN{Stephen Casper$^*$}
\IEEEauthorblockA{\textit{MIT CSAIL} \\
\textit{scasper@mit.edu}}
\and
\IEEEauthorblockN{Dylan Hadfield-Menell}
\IEEEauthorblockA{\textit{MIT CSAIL}}
}


\maketitle

\begingroup\renewcommand\thefootnote{}
\footnotetext{$*$ Equal contribution}

\begin{abstract}

The last decade of machine learning has seen drastic increases in scale and capabilities. Deep neural networks (DNNs) are increasingly being deployed in the real world. However, they are difficult to analyze, raising concerns about using them without a rigorous understanding of how they function. Effective tools for interpreting them will be important for building more trustworthy AI by helping to identify problems, fix bugs, and improve basic understanding. In particular, ``inner'' interpretability techniques, which focus on explaining the internal components of DNNs, are well-suited for developing a mechanistic understanding, guiding manual modifications, and reverse engineering solutions.

Much recent work has focused on DNN interpretability, and rapid progress has thus far made a thorough systematization of methods difficult. In this survey, we review over 300 works with a focus on inner interpretability tools. We introduce a taxonomy that classifies methods by what part of the network they help to explain (weights, neurons, subnetworks, or latent representations) and whether they are implemented during (intrinsic) or after (post hoc) training. To our knowledge, we are also the first to survey a number of connections between interpretability research and work in adversarial robustness, continual learning, modularity, network compression, and studying the human visual system. We discuss key challenges and argue that the status quo in interpretability research is largely unproductive. Finally, we highlight the importance of future work that emphasizes diagnostics, debugging, adversaries, and benchmarking in order to make interpretability tools more useful to engineers in practical applications.

\end{abstract}


\section{Introduction} \label{intro}

A defining feature of the last decade of deep learning is drastic increases in scale and capabilities \cite{sevilla2021parameters, kaplan2020scaling},
with the training compute for machine learning systems growing by ten orders of magnitude from 2010 to 2022 \cite{sevilla2022compute}. 
At the same time, deep neural networks (DNNs) are increasingly being deployed in the real world. 
If rapid progress continues, broad-domain artificial intelligence could be highly impactful \cite{bostrom2014superintelligence, muller2016future, tegmark2017life, russell2019human, christian2020alignment, ord2020precipice}.

Given this potential, it is important that practitioners can understand how AI systems make decisions, especially their issues. 
Models are most typically evaluated by their performance on a test set for a particular task.
This raises concerns because a black box performing well on a test set does not imply that the learned solution is adequate.
Testing sets typically fail to capture the full deployment distribution, including potential adversarial inputs.
They also fail to reveal problems with a model that do not directly relate to test performance (e.g. learning harmful biases). 
Moreover, even if a user is aware of inadequacies, the black-box nature of a system can make it difficult to fix issues.
Thus, a key step to building safe and trustworthy AI systems is to have an expanded toolbox for detecting and addressing problems.

We define an \textit{interpretability method} as any process by which an AI system's computations can be characterized in human-understandable terms.
This encompasses a broad set of techniques in the literature on DNNs, but in this paper, we focus specifically on methods for understanding internal structures and representations (i.e. not data, inputs, outputs, or the model as a whole).
We call these \emph{inner} interpretability methods.
We introduce a taxonomy for these methods, provide an overview of the literature, highlight key connections between interpretability and other topics in deep learning, and conclude with directions for continued work. 
Our central goals are twofold: (1) to provide a thorough resource for existing inner interpretability work and (2) to propose directions for continued research.



\begin{figure*}[t]
    \centering
    \includegraphics[
    width=0.8\textwidth]{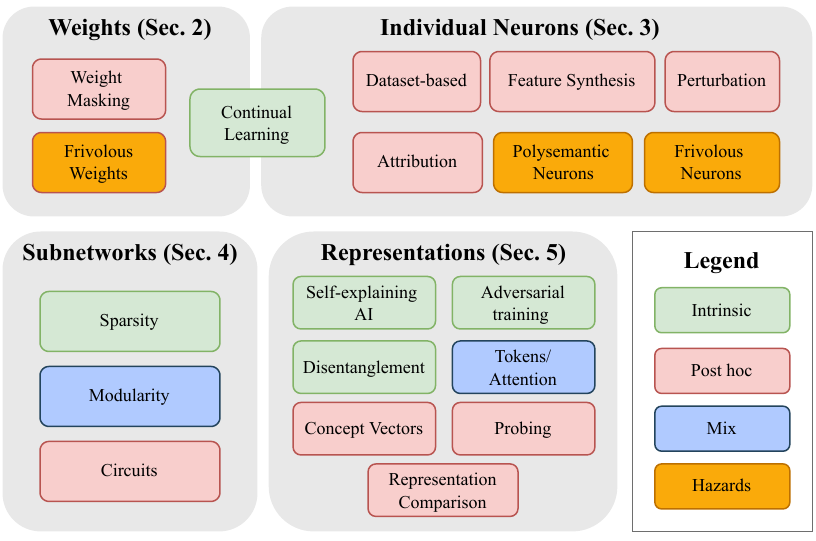}
    \caption{A taxonomy of inner interpretability methods and hazards associated with them. This mirrors our organization of Sections \ref{sec:weights}-\ref{sec:internal_representations}. We organize methods first by what part of the network's computational graph they help to explain: \textbf{weights}, \textbf{neurons}, \textbf{subnetworks}, or \textbf{latent representations}. Second, we organize approaches by whether they are \textbf{intrinsic} (implemented during training), \textbf{post hoc} (implemented after training), or can rely on a \textbf{mix} of intrinsic and post hoc techniques. Finally, we also discuss prominent \textbf{hazards} for different approaches.}
    \label{fig:taxonomy}
\end{figure*}

\subsection{The Importance of Interpretable AI}
\label{sec:importance}

\noindent Here, we outline several major motivations.

\medskip

\noindent \textbf{Open-ended evaluation:} Short of actually deploying a system, any method of evaluating its performance can fundamentally only be a proxy for its performance. 
In particular, test sets can fail to reveal--and often incentivize--undesirable solutions such as dataset bias, socially harmful biases, or developing deceptive solutions. 
Thus, it is important to have additional ways of rigorously evaluating systems' performance. 
One of the most important advantages of interpretability techniques lies in their unique ability to, unlike standard evaluation methods, allow humans to more open-endedly study a model and search for flaws.

\medskip

\noindent \textbf{Showcasing failure:} Uncovering why a model fails to produce a correct output can offer insights into what failures look like and how to detect them. 
This can help researchers avoid issues and help regulators establish appropriate rules for deployed systems.

\medskip

\noindent \textbf{Fixing bugs:} By understanding a problem and/or producing examples that exploit it, a network can be redesigned, probed, fine-tuned, and/or adversarially-trained to better align it with the user's goals.

\medskip

\noindent \textbf{Determining accountability:} Properly characterizing behavior is essential for establishing responsibility in the case of misuse or deployment failures.

\medskip

\noindent \textbf{Improvements in basic understanding:} By offering users more basic insights on models, data, and/or algorithms, interpretability techniques could be useful for reducing risks in deployed systems or better forecasting progress in AI. 
However, improved basic understanding could also be harmful if it causes advancements in risky capabilities to outpace effective oversight. 
We discuss this in Section \ref{sec:future_work}.
%

\medskip

\noindent \textbf{``Microscope'' AI:} Rigorously understanding how an AI system accomplishes a task may provide additional domain knowledge.
This could include insights about solving the task as a whole or the properties of specific examples.
This goal has been referred to as ``microscope'' AI \cite{hubinger2020overview}, and it could allow for reverse engineering more understandable or verifiable solutions.
This may be especially valuable for studying systems with superhuman performance.

\begin{figure*}[t]
    \centering
    \includegraphics[width=0.55\textwidth]{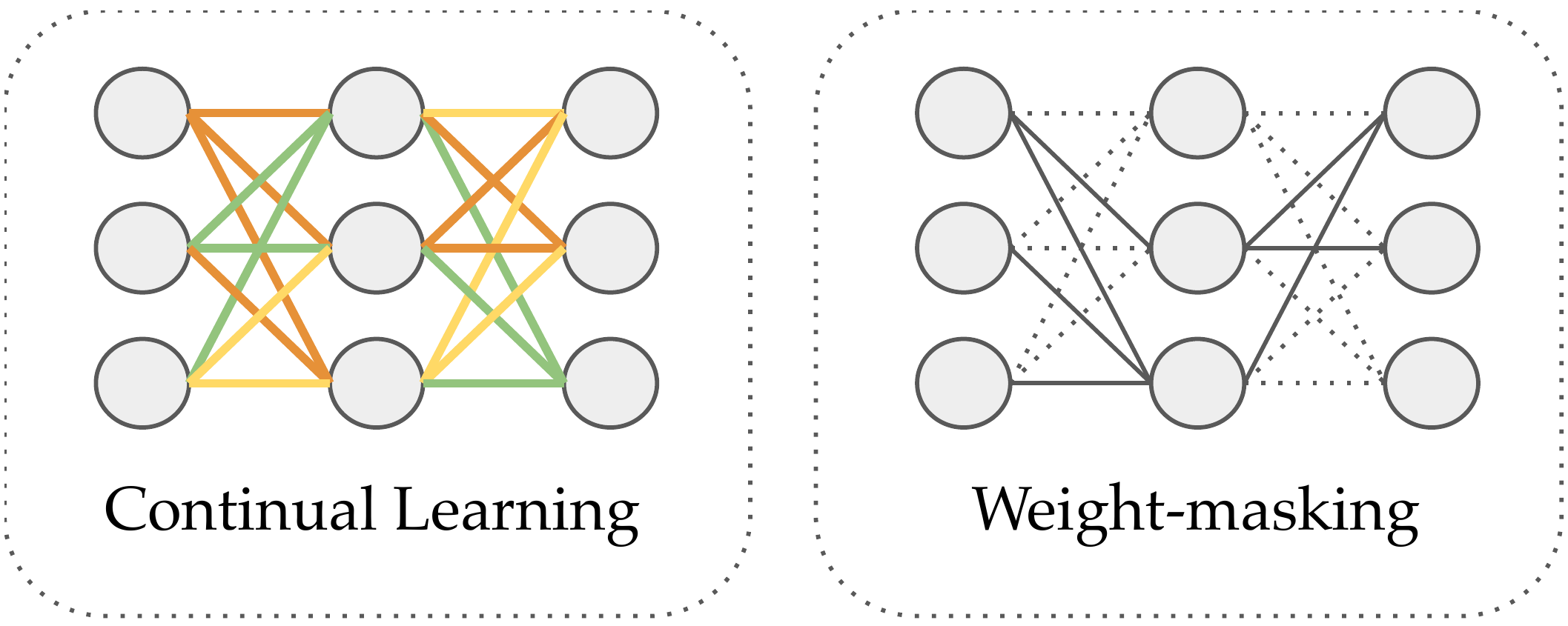}
    \caption{Inner interpretability methods for weights can focus on (1) \textbf{continual learning} techniques that make weights specialize in particular tasks or (2) \textbf{weight-masking} techniques which learn a mask over weights as a way of discovering which weights are key for a certain task.}
    \label{fig:neuron_methods}
\end{figure*}

\subsection{Scope}

\noindent \textbf{Inner Interpretability:} Our focus is on \emph{inner} interpretability methods for DNNs. 
Black-box techniques, adversarial techniques, input attribution methods, neurosymbolic methods, and ``good old-fashioned AI'' are all valuable but beyond the scope of this survey.
This is not to say that they are of less value to building safer AI than the methods we focus on -- many of them have major advantages, and a diverse interpretability toolbox is important. 
However, we focus on inner interpretability methods because (1) there is a great deal of current interest in them and (2) they are well-equipped for certain goals such as guiding manual modifications, reverse engineering solutions, and detecting inner ``latent'' knowledge which may contribute to deceptive behavior.

\medskip

\noindent \textbf{Contrasts with past survey works:} See also several previous surveys and critiques of interpretability work that overlap with ours \cite{doshivelez2017towards, lipton2018mythos, adadi2018peeking, zhang2018visual, rudin2019stop, miller2019explanation, samek2019towards, gilpin2018explaining, molnar2020interpretable, danilevsky2020survey, jacovi2020towards, molnar2020pitfalls, das2020opportunities, krishnan2020against, samek2021explaining, sajjad2021neuron, rudin2022interpretable, molnar2022}.
Unlike this survey, \cite{doshivelez2017towards, krishnan2020against, lipton2018mythos, rudin2019stop, miller2019explanation} are critique/position papers that do not extensively survey existing work, \cite{adadi2018peeking, das2020opportunities, gilpin2018explaining, samek2019towards, molnar2020interpretable, molnar2020pitfalls, samek2021explaining, rudin2022interpretable} focus mostly or entirely on approaches out of the scope of this work (e.g. non-DNNs or feature attribution), \cite{danilevsky2020survey, jacovi2020towards} only survey methods for language models, \cite{zhang2018visual} only focuses on convolutional networks, \cite{sajjad2021neuron} only surveys single-neuron methods, \cite{molnar2020interpretable, samek2021explaining} only focus on post-hoc methods, and \cite{doshivelez2017towards, zhang2018visual, lipton2018mythos, adadi2018peeking, samek2019towards, rudin2019stop, miller2019explanation, gilpin2018explaining} are relatively old (before 2020).
This survey is also distinct from all of the above in its focus on inner interpretability, implications for diagnostics and debugging, and the intersections between interpretability and a number of other research paradigms. 

\subsection{Taxonomy}

Our taxonomy divides inner interpretability techniques by what part of the DNN's computational graph they explain: \textbf{weights}, \textbf{neurons}, \textbf{subnetworks}, or \textbf{latent representations}.
We dedicate Sections \ref{sec:weights}-\ref{sec:internal_representations} respectively to each of these approaches.
Interpretability techniques can also be divided by whether they are used during or after training.
\textbf{Intrinsic} interpretability techniques involve training models to be easier to study or come with natural interpretations. 
\textbf{Post hoc} techniques aim to interpret a model after it has been trained.
We divide methods by whether they are intrinsic or post hoc at the subsection level.

Fig. \ref{fig:taxonomy} depicts our taxonomy and previews the organization of Sections \ref{sec:weights}-\ref{sec:internal_representations}.
Note that this taxonomy sometimes divides related methods. 
For example, continual learning methods for weights (Section \ref{subsec:continual_learning_weights}) and neurons (Section \ref{subsec:continual_learning_neurons}) are conceptually similar, and methods for interpreting subnetworks (Section \ref{sec:subnetworks}) frequently involve variations or applications of methods for weights \ref{sec:weights}) or neurons \ref{sec:individual_neurons}).
We note these connections as we discuss the families of methods below. 
However, we divide methods first by what part of the network they target because how a technique \emph{operates} on a network typically matters more for goal-oriented engineering than whether it occurs during or after training.

\section{Weights} \label{sec:weights}


\subsection{Continual Learning (Intrinsic):} \label{subsec:continual_learning_weights}

One research paradigm in deep learning is to train systems that can learn new tasks without forgetting old ones. 
This is known as \emph{continual learning} or avoiding \emph{catastrophic forgetting} \cite{de2019bias, smith2022closer}.
Some techniques are based on the principle of having weights specialize for particular types of input data, updating more for some than others \cite{kirkpatrick2017overcoming, li2017learning, zenke2017continual, mallya2018packnet, aljundi2019task, 
ahn2019uncertainty, 
titsias2019functional}. 
This offers a natural way to characterize weights based on the tasks or classes that they specialize in.
Unfortunately, current research on these methods has not been done with an emphasis on improving interpretations of weights or subnetworks.
This may be a useful direction for future work. 
See also methods for continual learning that operate on neurons in Section \ref{subsec:continual_learning_neurons}.

\subsection{Weight-Masking (Post Hoc):} \label{subsec:weight_masking}
In contrast to intrinsic methods, one can also train weight masks over a network to determine which weights are essential for which tasks \cite{csordas2020neural, wortsman2020supermasks, zhao2020masking}. 
For example, a mask over a classifier's weights can be trained to cover as many as possible while preserving performance on a subset of data. 
The resulting mask identifies a subset of weights (and a corresponding subnetwork) that can be causally understood as specializing in that subtask.
This approach also works for identifying subnetworks that specialize in a task (Section \ref{sec:subnetworks}).

\subsection{Frivolous Weights (Hazard):} \label{subsec:frivolous_weights}

A difficulty in interpreting weights is that many are often unimportant to the network. 
Past works have shown that networks can often be pruned to contain a very small fraction of their original weights with little to no loss in performance (though sometimes with fine-tuning) \cite{frankle2018lottery, blalock2020state, vadera2020methods}.
See also frivolous neurons (Section \ref{subsec:frivolous neurons}).

\begin{figure*}[t]
    \centering
    \includegraphics[width=0.85\textwidth]{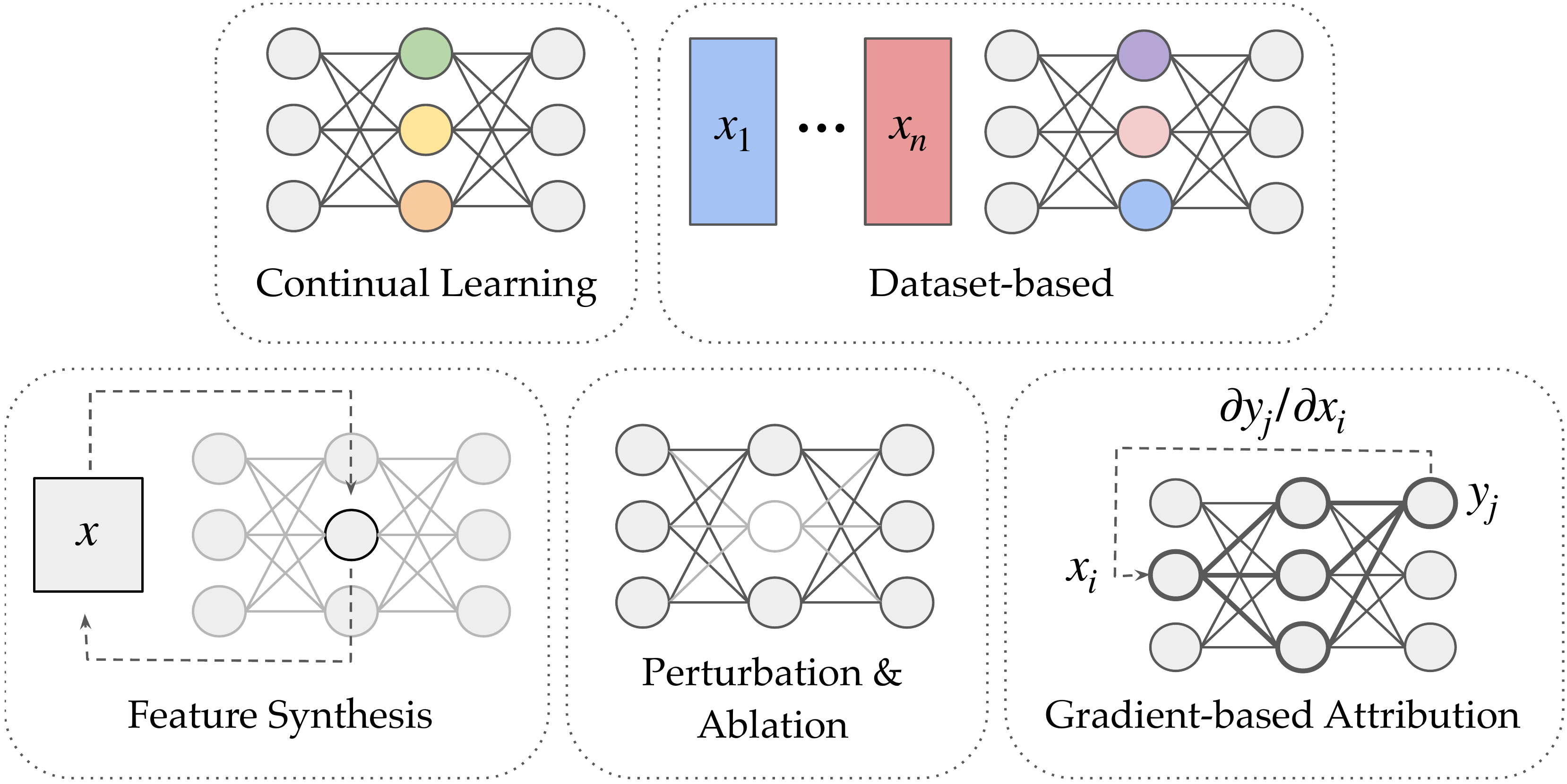}
    \caption{Inner interpretability methods for individual neurons can focus on (1) \textbf{continual learning} techniques that make neurons specialize in particular tasks, (2) \textbf{dataset-based} techniques that aim to find which neurons respond to which features, (3) \textbf{feature synthesis} to construct inputs that excite individual neurons, (4) \textbf{perturbation or ablation} of neurons coupled with analysis of changes to network behavior, and (5) \textbf{gradient-based attribution} methods that analyze partial derivatives of outputs w.r.t. neural activations.}
    \label{fig:neuron_methods}
\end{figure*}

\section{Individual Neurons} \label{sec:individual_neurons}

\noindent As is common in the literature, we use ``neuron'' to refer both to units in dense layers and elements of feature maps in convolutional layers.

\subsection{Continual Learning (Intrinsic):} \label{subsec:continual_learning_neurons}

Just as continual learning \cite{de2019bias, smith2022closer} can be facilitated via specialization among weights (see Section \ref{subsec:continual_learning_weights}), the same can be done with neurons. 
Unlike weight-based continual learning methods, which have weights update more for some tasks than others, neuron-based ones typically rely on adding new neurons to the architecture upon encountering a new task \cite{rusu2016progressive, yoon2017lifelong,
lee2020neural}. 
This discourages neurons from learning to simultaneously detect features from multiple unrelated tasks and allows for natural interpretations of neurons in terms of what subtasks they specialize in. 
As with continual learning methods that operate on weights, current research on these methods has not been done with an emphasis on improving interpretations of neurons or subnetworks.
This may be a useful direction for future work. 
See also Section \ref{subsec:modularity}, which discusses methods for modularity among neurons.

\subsection{Dataset-Based (Post Hoc):} \label{subsec:dataset_based}

A simple way to characterize the role of individual neurons is to use a dataset to analyze which types of inputs they respond to. 
Perhaps the simplest example of this is searching through a dataset and selecting the inputs that maximally excite a given neuron \cite{zhou2014object}. 
A more sophisticated technique known as network ``dissection'' uses a richly-labeled dataset of semantic concepts to analyze neural responses \cite{bau2017dissection, bau2018GAN, Bau2020understanding}. 
A neuron can then be characterized based on how well its activations align with different types of input. 
This line of work has been extended to assign descriptions to neurons using compositional logic expressions on a set of labels \cite{mu2020compositional}. 
This allows the interpretability of neurons to be quantified as the intersection over union for a logical formula on input features and the neuron's activations.
This has been further extended to develop natural language explanations by using captioning methods to describe a set of image patches that activate a neuron \cite{hernandez2021natural, oikarinen2022clip_dissect}.
These approaches have proven useful for identifying undesirable biases in networks \cite{mu2020compositional, hernandez2021natural}.
Dissection has also been used to analyze what types of neurons are exploited by adversarial examples \cite{xu2019interpreting}, identify failure modes for text-to-image models \cite{cho2022dall}, and probe neural responses in transformers to isolate where specific information is stored \cite{vig2020investigating, finlayson2021causal, geiger2021causal, dai2021knowledge, geiger2022inducing, meng2022locating}.
This can then be followed by improving the model by editing a learned fact (such as an undesirable bias) \cite{dai2021knowledge, meng2022locating}.
Unfortunately, all dataset-based methods are limited by the diversity of examples in the dataset used and the quality of labels. 
See also probing methods in Section \ref{subsec:probing}.

\subsection{Feature Synthesis (Post Hoc):} \label{subsec:feature_synthesis}

This approach is based on synthesizing inputs with the goal of maximally (or minimally) activating a given neuron or combination of neurons.
Synthesis methods come with the advantage of not being limited to a particular dataset.
Several works have taken this approach, optimizing inputs to excite particular neurons \cite{
mahendran2015understanding, 
nguyen2016multifaceted, 
olah2017feature}. 
One can use a distance measure in the optimization objective to synthesize a batch of inputs to be diverse \cite{olah2017feature}. 
There has also been work on using generative models instead of directly optimizing over input features \cite{nguyen2016synthesizing, nguyen2016plug, casper2021robust, casper2022diagnostics}.
A broader survey of these types of methods is provided by \cite{nguyen2019understanding}.
However \cite{borowski2020exemplary} finds that natural exemplars which strongly-activate individual neurons can be more useful for helping humans predict neural responses to data than synthesized features.

\subsection{Neural Perturbation and Ablation (Post Hoc):} \label{subsec:neural_perturbation}

By analyzing a DNN's behavior under perturbation to a neuron, one can gain insight into the type of information it processes.
For example, if a neuron in an image classifier robustly and uniquely detects dogs, one should expect performance on dog classification to worsen when that neuron is ablated (i.e. dropped out). 
A key benefit of these methods is that they allow for testing counterfactuals, helping establish a causal rather than a correlational relationship between neural activations and the behavior of the network. 
Works in this area have used neural ablation \cite{zhou2018revisiting, hod2021detecting}, subspace ablation \cite{morcos2018importance, ravfogel2022linear}, and non-ablation perturbations \cite{bau2018identifying, elhage2021mathematical}.
Notably, the net effect of perturbing a neuron can vary by context and which others, if any, are also perturbed.
To account for this, one can compute Shapley values for neurons to measure their importance averaged over the ablation of other neurons \cite{Stier2018analysing, ghorbani2020shapley}, and this has been shown to be a practical way of identifying neurons that can be removed or modified to reduce bias or improve robustness \cite{ghorbani2020shapley}.
Shapley values, however, are limited in their ability to provide useful causal explanations \cite{kumar2020problems}.

\subsection{Gradient-Based Attribution (Post Hoc):} \label{subsec:gradient_based_attribution}

Much work has been done on gradient-based feature attribution to study which features are influential for neural responses or model outputs.
There are several surveys and critiques of feature attribution methods in particular \cite{dombrowski2019explanations, adebayo2018sanity, zhang2019should, ancona2019gradient, slack2020fooling, jeyakumar2020can, adebayo2020debugging, nielsen2021robust, denain2022auditing, fokkema2022attribution, holmberg2022towards, jyoti2022robustness}.
Most of this work has been done to study attributions on \emph{inputs} and is outside the scope of this survey. 
However, the same type of approach has been applied for attribution with internal neurons. 
\cite{sundararajan2017axiomatic} introduce an approach for this using gradients along with runtime tests for sensitivity and invariance to evaluate the quality of interpretations.  
Building off this, several works have found gradient-based attribution useful in large language models \cite{ancona2017towards, durrani2020analyzing, lundstrom2022rigorous}, including to guide a search for where certain facts are stored \cite{dai2021knowledge}.
However, these methods are limited in that explanations are only as valid as the local linear approximation the gradient is based on, and they cannot directly provide causal explanations.

\subsection{Polysemantic Neurons (Hazard):}  \label{subsec:polysemantic_neurons}

\emph{Polysemantic} neurons are activated by multiple unrelated features.
They have been discovered via dataset-based methods \cite{fong2018net2vec, mu2020compositional, hernandez2021natural}, various forms of visual feature synthesis \cite{nguyen2016multifaceted, olah2020zoom, voss2021visualizing, goh2021multimodal}, and feature attribution \cite{elhage2022solu}. 
How and why they form remains an open question.
However, \cite{olah2020zoom} observed a tendency for monosemantic neurons to become polysemantic over the course of training and hypothesized that it is associated with representing information more efficiently.
This would suggest that polysemantic neurons might be useful for model performance. 
However, they also pose a significant challenge for two reasons.
First, interpretations of polysemantic neurons are more likely to be incorrect or incomplete.
Second, it has been shown that they can be exploited for adversarial attacks \cite{mu2020compositional, hernandez2021natural}.
See also Section \ref{subsec:disentanglement} for a discussion of \emph{entanglement}, which generalizes the notion of polysemanticity to layers.

\begin{figure*}[t]
    \centering
    \includegraphics[width=0.85\textwidth]{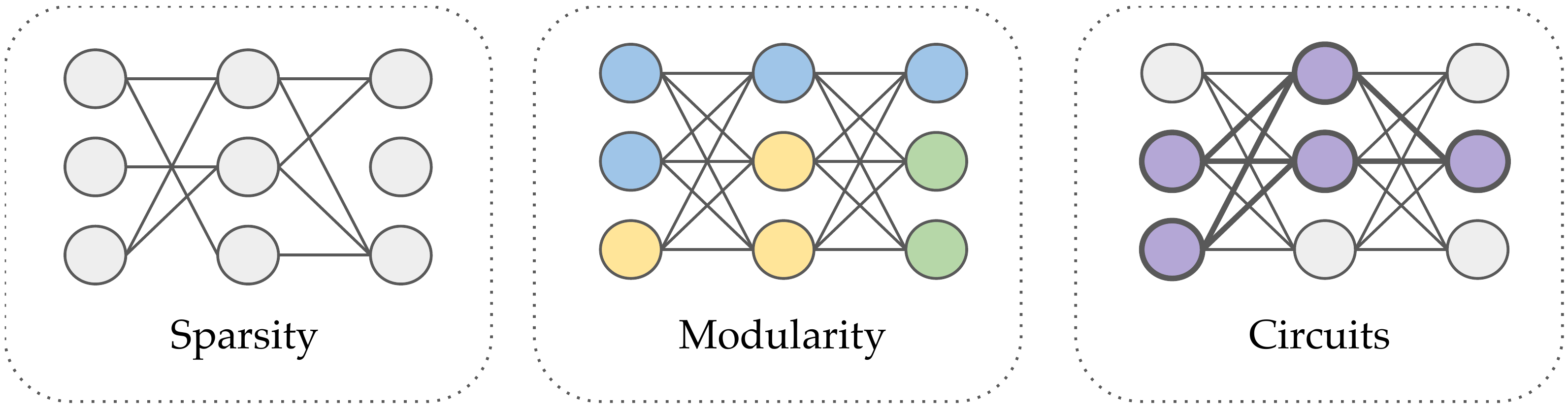}
    \caption{Inner interpretability methods for subnetworks can focus on (1) simplifying the computational subgraph via \textbf{sparsity}, (2) either intrinsic methods to enforce \textbf{modularity} among neurons or post hoc methods to group them into modules, or (3) analysis of neural \textbf{circuits} which can be understood as performing a specific task.}
    \label{fig:subnetworks_methods}
\end{figure*}

\subsection{Frivolous Neurons (Hazard):} \label{subsec:frivolous neurons}

Frivolous neurons are not important to a network. 
\cite{casper2019frivolous} defines and detects two distinct types: \emph{prunable} neurons, which can be removed from a network by ablation, and \emph{redundant} neurons, which can be removed by refactoring weight matrices. 
They pose a challenge for interpretability because a frivolous neuron's contribution to the network may be meaningless or difficult to detect with certain methods (e.g., neural perturbation). 
Network compression may offer a solution.
For example, \cite{sainath2013low, srinivas2015data, hu2016network, luo2017thinet, 
he2020learning} each compress networks by eliminating frivolous neurons. 
Compression and the interpretability of neurons are linked. 
After compressing a network, \cite{li2019exploiting} found that the remaining neurons were more interpretable with only marginal change in performance, and \cite{yao2021deep} used proxies for neuron interpretability to guide neuron-level pruning. 
Additionally, the motivation for pruning to increase interpretability is closely-related to intrinsically interpretable layer representations. 
See also Section \ref{subsec:frivolous_weights} on frivolous weights and Section \ref{subsec:disentanglement} on neural disentanglement.

\section{Subnetworks} \label{sec:subnetworks} 

Note that many of the methods used for analyzing subnetworks supervene on techniques for weights (Section \ref{sec:weights}) or neurons (Section \ref{sec:individual_neurons}). 

\subsection{Sparsity (Intrinsic):} \label{subsec:sparsity}

Sparse weights inside of DNNs allow for much simpler analysis of relationships between neurons. 
In some cases, sparsification can reduce the number of weights by almost two orders of magnitude while causing little to no tradeoff with performance \cite{frankle2018lottery}.
Sparsity-aided interpretability has been explored through pruning \cite{
frankle2019dissecting, wong2021leveraging, bena2021extreme, moran2021identifiable} regularization \cite{ross2017neural}, and sparse attention \cite{meister2021sparse}.
In particular, \cite{wong2021leveraging} demonstrates how sparsity can be paired with post-hoc techniques for neuron analysis to help a human to edit a model.
This has direct implications for safety and debiasing.
Pruning portions of the network architecture can also be guided by measures of interpretability \cite{yeom2021pruning, wang2020dynamic}. 
Meanwhile, as an alternative to conventional sparsity, \cite{wu2018beyond} introduce a method to regularize the behavior of a neural network to mimic that of a decision tree.

While sparsity simplifies the analysis of subnetworks, it may not improve the interpretability of \emph{individual neurons}. 
\cite{frankle2019dissecting} find no increase in their interpretability through the dissection of pruned networks, and \cite{meister2021sparse} fail to find evidence of improved interpretability of individual neurons with sparse attention.

\subsection{Modularity (Intrinsic):}  \label{subsec:modularity}

Modularity is a common principle of engineered systems and allows for a model to be understood by analyzing its parts separately.
At a high level, \cite{amer2019review} offers a survey of DNN modularization techniques, and \cite{agarwala2021one, mittal2022modular} study the capabilities and generality of modular networks compared to monolithic ones.
The simplest way to design a modular DNN is to use an explicitly modular architecture. 
This can be considered ``hard'' modularity. 
This can be a form of ``model-aided deep learning'' \cite{shlezinger2020model} if domain-specific considerations are used to guide the design. 
Modular architectures were studied by \cite{voss2021branch} who analyzed the extent to which neurons in a branched architecture learned to process different features from those in other branches, and \cite{yan2020neural} who experimented with distinct neural modules that were trained to execute algorithmic subroutines.

Aside from branched architectures, a ``softer'' form of modularity can be achieved if neurons in different modules are connected to each other but must compete for access to information. 
This can allow for end-to-end differentiability, yet sparse information flow between modules. 
Methods for soft modularity have been studied via initialization \cite{filan2021clusterability}, regularization \cite{filan2021clusterability}, a controller \cite{kirsch2018modular, jiang2019self}, or sparse attention \cite{andreas2016neural, goyal2019recurrent, serra2018overcoming, elhage2022solu}.
Notably \cite{serra2018overcoming} used attention to both induce specialization among neurons and reduce catastrophic forgetting. 
See also methods for avoiding catastrophic forgetting by having subsets of neurons specialize in a given task in Section \ref{subsec:continual_learning_neurons}.

\subsection{Modular Partitionings (Post Hoc):}  \label{subsec:modular_partitionings}

A post hoc way of understanding a DNN in terms of modules is to partition the neurons into a set of subnetworks, each composed of related neurons. 
Toward this goal, \cite{watanabe2018modular, watanabe2019understanding, filan2021clusterability} divide neurons into modules based on graphical analysis of the network's weights and analyze how distinct the neurons in each module are.
These methods involve no data or runtime analysis. 
In contrast, \cite{watanabe2019interpreting, arik2020explaining, hod2021detecting, casper2022graphical, lange2022clustering} each perform partitioning and cluster analysis based on how neurons associate with inputs and/or outputs.
In particular, \cite{hod2021detecting} present a statistical pipeline for estimating the interpretability of neuron clusters without a human in the loop.
Overall, however, these methods have had very limited success in finding highly-composite partitionings in models. 
A useful direction for future work may be to combine this approach with intrinsic modularity methods.

\subsection{Circuits Analysis (Post Hoc):} \label{subsec:circuits_analysis}

Instead of analyzing an entire partition of a network, a much simpler approach is to study individual subnetworks inside of it. 
These have often been referred to as neural ``circuits'' which can be as small as just a few neurons and weights. 
This has been done with weight masking \cite{wang2018interpret, csordas2020neural}, data-based methods \cite{zhang2018explanatory, fiacco2019deep, townsend2020eric, zhang2020extraction, santurkar2021editing}, feature synthesis \cite{olah2017feature, olah2018building, olah2020overview, cammarata2020curve, olah2020naturally, olah2020zoom, voss2021visualizing, schubert2021high, petrov2021weight}, and neural ablation \cite{meyes2020under, hamblin2022pruning}.
However, many of the successes of circuits analysis to date have focused on toy models and involved intensive effort from human experts. 
To be useful for improving models in practical applications, future methods will likely need to leverage automation. 
\cite{ren2021towards} make progress on this by distilling a DNN into a sparse symbolic causal graph on a set of concepts and providing theoretical guarantees for the faithfulness of the graph.
See also Section \ref{subsec:tokens_and_attention} for a discussion of circuits in transformers. 

\begin{figure*}[t]
    \centering
    \includegraphics[width=0.95\textwidth]{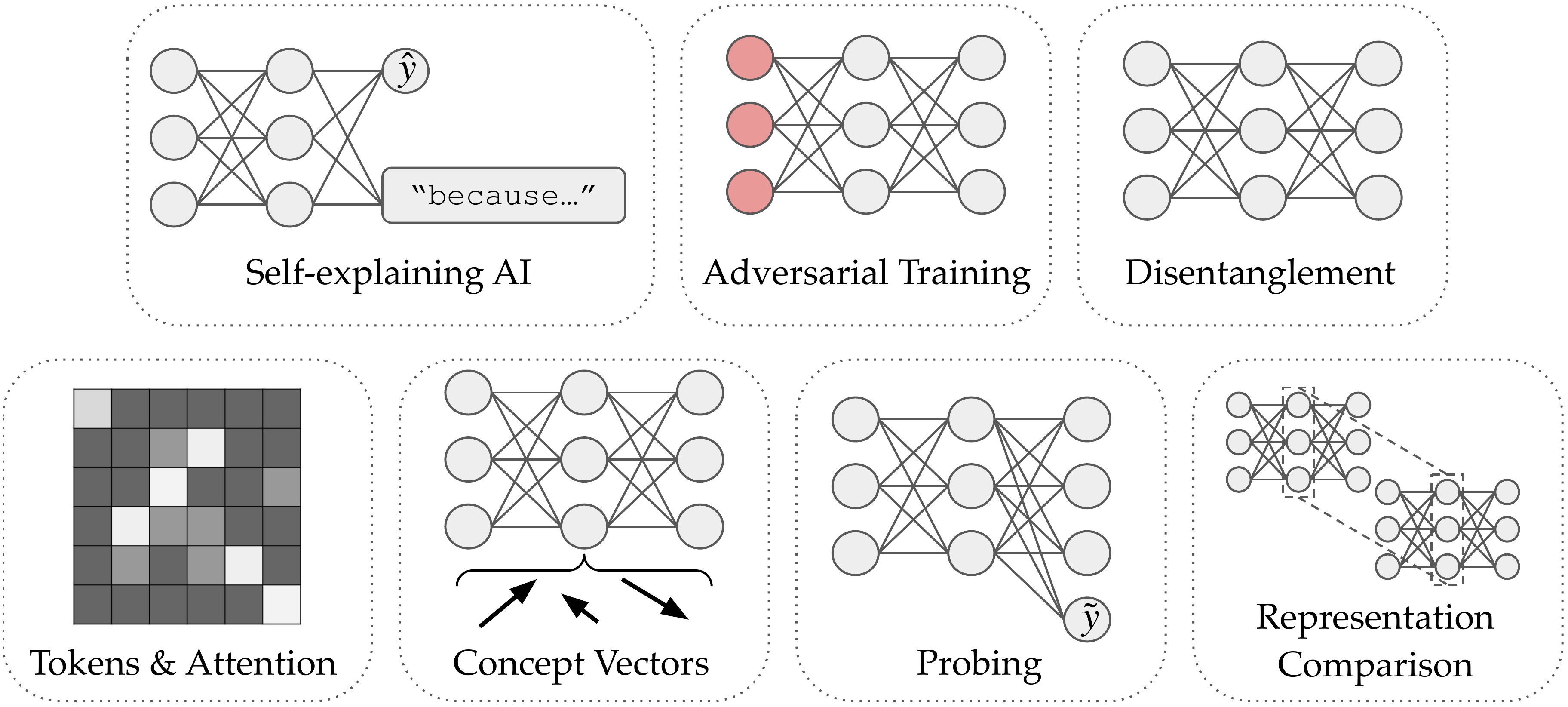}
    \caption{Inner interpretability methods for neural representations can focus on (1) training \textbf{self-explaining AI} systems that explain their decisions, (2) \textbf{adversarial training}, (3) the \textbf{disentanglement} of latent representations such that each neuron tends to uniquely respond to a single concept in data, (4) analysis of \textbf{token} evolution or \textbf{attention} maps in transformers, (5) analysis of \textbf{concept vectors} in latent space, (6) \textbf{probing} neural representations to evaluate their transferability to a target task, and (7) \textbf{representation comparison} between different layers in two networks.}
    \label{fig:representations}
\end{figure*}

\section{Internal Representations} \label{sec:internal_representations}

\subsection{Self-Explaining Models (Intrinsic):} \label{subsec:self_explaining_models}

Most methods in the literature used for understanding DNNs aim to help a human ``open up'' the network and study parts of it. 
If one wants to understand another human's reasoning, the analogous techniques would involve studying their brain directly. 
These are sometimes useful, but in most cases, simply asking another human for an explanation of what they are thinking is much more effective. 
Self-explaining AI systems are meant to provide such explanations of internal reasoning in an analogous way to how humans can provide them.
Competing definitions are offered in the literature, but we will use one based on \cite{elton2020self}, which simply requires that a model produces an explanation for its reasoning that can be easily be understood by a human, ideally paired with a confidence estimate.

In computer vision, one approach has been to classify images based on their similarity to a set of learned ``prototypes'' \cite{kim2014bayesian, li2018deep, alvarez2018towards, chen2019looks, rymarczyk2021interpretable, zhang2021hypothesis}.
Prototype-based classification has also been studied in language models \cite{Farhangi_2022}.
These methods allow the model to attribute its outputs to a set of exemplary datapoints, allowing its decision to be explained as ``\emph{this} input resembles \emph{these} other examples.''

Another self-explaining AI strategy has been to supervise human-understandable explanations for model outputs that are computed from the same inner representations. 
In computer vision, this has been done for classification and question answering \cite{hendricks2016generating, hendricks2018grounding, kim2018textual, akata2018generating, patro2020robust}. 
In natural language processing, this has been done for question answering and natural language inference with explanations \cite{camburu2018snli, lamm2020qed, kumar2020nile, zhao2020lirex}.
For large language models that have sufficiently general language capabilities, explanations can also simply be elicited via prompts (e.g., \cite{brown2020language, chowdhery2022palm}).
However, the extent to which these explanations accurately explain the model's decision making is very unclear \cite{kadavath2022language}.

\cite{alvarez2018towards} argues that explanations should meet three criteria: (1) Explicitness: are explanations direct and understandable? (2) Faithfulness: do they correctly characterize the decision? And (3) Stability: how consistent are they for similar examples?
It has been shown that explanations from such models can be unfaithful \cite{alvarez2018towards, valentino2021natural} or vulnerable to adversarial examples \cite{zheng2019analyzing, camburu2019make, hoffmann2021looks}, so producing self-explaining models that meet these remains an open challenge. 
Toward fixing these issues, \cite{deyoung2019eraser} introduces an NLP benchmark, and \cite{bontempelli2021toward} provides an interactive debugging method for prototype networks.

\subsection{Adversarial Training (Intrinsic)} \label{subsec:adversarial_training}

\cite{engstrom2019adversarial} found that adversarially trained classifiers exhibited improvements in a number of interpretability-related properties, including feature synthesis for neurons (see Section \ref{subsec:feature_synthesis}).
It has also been found that these adversarially trained networks produce better representations for transfer learning \cite{salman2020adversarially}, image synthesis \cite{santurkar2019image, casper2021robust, casper2022diagnostics}, and for modeling the human visual system \cite{engstrom2019learning}.
Unfortunately, robustness may be at odds with accuracy \cite{tsipras2018robustness}, potentially due to predictive but ``nonrobust'' features in a dataset \cite{ilyas2019adversarial}. 
This had led to an understanding that adversarial examples can be used to help to understand what types of useful or exploitable features a network detects and represents \cite{casper2021robust, casper2022diagnostics}.

\subsection{Disentanglement (Intrinsic):} \label{subsec:disentanglement}

During a pass through a network, each layer's activations can be represented as a vector in latent space.
The goal of \emph{disentanglement} \cite{bengio2013representation} is ensuring that features can be more easily identified from studying a latent vector by encouraging a more bijective relationship between neurons and a set of interpretable concepts. 
See also Section \ref{subsec:polysemantic_neurons} for a discussion of polysemantic neurons. 
Disentanglement can be done in a supervised manner by encouraging neurons to align to a set of predetermined concepts.
\cite{chen2020concept} did this by applying a whitening operation to decorrelate features followed by a learned orthogonal transformation to produce latent activations that could be supervised.
Similarly, inner supervision was used by \cite{losch2019interpretability, koh2020concept, losch2021semantic} to train a `bottleneck' layer to separate features, and by \cite{subramanian2018spine} to learn sparse, interpretable embeddings.
However, \cite{mahinpei2021promises} discusses challenges with these models, particularly the problem of ``leakage'' in which undesired information nonetheless makes it through the bottleneck.

Disentanglement can also be done in an unsupervised manner. 
A partial example of this is dropout \cite{JMLR:v15:srivastava14a} which prevents co-adaptation among neurons, though at the cost of increasing redundancy.
Other works have explored using lateral inhibition between neurons in a layer to make them compete for activation \cite{cao2018lateral, krotov2019unsupervised, subramanian2018spine, elhage2022solu}, designing a ``capsule'' based architecture in which a group of neurons have activations that each represent a specific feature \cite{sabour2017dynamic, deliege2019effective}, aligning activations to components of variation in data \cite{kuo2019interpretable}, using a mutual information loss \cite{chen2016infogan}, using an inter-class activation entropy-based loss \cite{zhang2018interpretable}, regularizing the Hessian of the network's outputs w.r.t a layer \cite{peebles2020hessian}, training a classifier and autoencoder from the same latents \cite{schneider2021explaining}, or learning a mask over features \cite{he2022exploring}.
Other works have focused specifically on autoencoders, training them to have more independently-activated neurons \cite{higgins2016beta, kumar2017variational, burgess2018understanding, kim2018disentangling, chen2018isolating}.
However, in a survey of these methods \cite{locatello2019challenging, locatello2020sober} prove an impossibility result for unsupervised disentanglement without inductive biases on both the models and data.

\subsection{Tokens and Attention (Intrinsic and Post Hoc):} \label{subsec:tokens_and_attention}

Transformer architectures process data by passing token representations through attention and feed forward layers in alternating fashion. 
These architectural building blocks pose unique opportunities for studying the network's internal representations.

First, the tokens can be studied.
This can be done by interpreting token representations in transformers directly \cite{li2021implicit, geva2022lm, geva2022transformer, elhage2021mathematical, olsson2022context, nanda2022mechanistic} or analyzing how fully-connected layers process them \cite{geva2020transformer, nanda2022mechanistic}.

Second, key-query products are computed inside of an attention layer and represent how much each inner token is attending to others. 
This notion of studying relations between token representations has similarities to circuits analysis covered in Section \ref{subsec:circuits_analysis}.
In their seminal work, \cite{bahdanau2014neural} showed that an attentional alignment appeared to show the expected attention patterns for translation. 
Other recent works have used this approach more systematically \cite{vashishth2019attention, clark2019does, hao2021self, abnar2020quantifying} including for the study of harmful biases \cite{de2019bias}. 
And \cite{rigotti2021attention} introduced a ``ConceptTransformer'' whose outputs can be explained as an attention map over user-defined concepts much like a concept bottleneck network \cite{koh2020concept}. 
Interactive tools for visual analysis of attentional attribution are provided by \cite{lee2017interactive, liu2018visual, strobelt2018s, vig2019multiscale}.
And \cite{elhage2021mathematical, chefer2021transformer, olsson2022context} expanded on this approach toward the goal of multi-step attribution across multiple layers. 
Importantly, the analysis of attention may not always suggest faithful explanations, and an over-reliance on them for interpretation can be hazardous \cite{jain2019attention, serrano2019attention, wiegreffe2019attention}.
Finally, transformers may have many frivolous, prunable attention heads \cite{voita2019analyzing}, suggesting a further need for caution because not all heads may be worth interpreting.

\subsection{Concept Vectors (Post Hoc):} \label{subsec:concept_vectors}

While disentanglement aims to align concepts with individual neurons, methods for analyzing concept vectors are post hoc solutions to the same problem. 
Here, the goal is to associate directions in latent space with meaningful concepts. 
Several works have done this by analysis of activations induced by images from a dataset of concepts \cite{fong2018net2vec, kim2018interpretability, zhou2018interpretable, reif2019visualizing, lucieri2020explaining, lucieri2020interpretability} including \cite{abid2022meaningfully, yuksekgonul2022post} who used it explicitly for debugging.
A contrasting approach was used by \cite{schneider2021explaining}. 
Rather than beginning with concepts and then identifying directions for them, they first identified directions using a generator and a ``layer selectivity'' heuristic, and then sought to find post hoc explanations of what they encoded.
A debugging-oriented approach was taken by \cite{jain2022distilling, wiles2022discovering} who classified and clustered embeddings of data examples that were incorrectly labeled by a classifier, including cases due to demographic biases. 
This allowed for detection, interpretation, and intervention for potentially difficult inputs for the model as well as a way to identify underrepresented subcategories of data. 
Unfortunately for these approaches, there is evidence that networks learn to represent many more useful concepts than can linearly independently be represented by their internal layers \cite{elhage2022toy}. 
And single directions in activation space can correspond to unrelated concepts depending on the activation vector's magnitude \cite{black2022interpreting}.

\subsection{Probing (Post Hoc):} \label{subsec:probing}
Given some way of embedding data, the goal of probing is to understand whether or not that embedding captures a certain type of information.
Probing leverages transfer learning as a test for whether embeddings carry information about a target task.
The three steps to probing are to (1) obtain a dataset that contains examples capturing variation in some quality of interest, (2) embed the examples, and (3) train a model on those embeddings to see if it can learn the quality of interest.
Any inner representation from any model can be used, making this a versatile technique.
A survey of probing works is provided by \cite{belinkov2022probing}.
The simplest example of probing is to use an unsupervised learning algorithm as the probe \cite{hoyt2021probing}. 
Additional work has been done with linear probes for image classifiers \cite{alain2016understanding}. 
However, probing has most commonly been done in language models \cite{gupta2015distributional, kohn2015s, adi2016fine, ettinger2016probing, conneau2018you, perone2018evaluation, niven2019probing, 
saleh2020probing, lepori2020picking, tamkin2020investigating, miaschi2021probing, lindstrom2021probing, li2022probing}.
Notably, a form of contrastive probing was used by \cite{burns2022discovering} for detecting deception in language models.
While versatile, probing is imperfect \cite{antverg2021pitfalls}.
One issue is that a probe's failure to learn to represent the desired quality in data is not necessarily an indicator that it is not represented. 
For example, this may be the case with an underparameterized probe. 
On the other hand, a successful probe does not necessarily imply that the model being probed actually uses that information about the data. 
This was demonstrated by \cite{ravichander2020probing} who argued for the use of rigorous controls when probing.
In a subsequent paper, \cite{elazar2021amnesic} aimed to address this problem by pairing probing with experiments that manipulated the data in order to analyze the causal influence of perturbations on performance.
See also dataset-based methods for characterizing neurons in Section \ref{subsec:dataset_based}.

\subsection{Representation Comparison (Post Hoc):} \label{subsec:measuring_representational_similarity}
A somewhat indirect way of characterizing the representations learned by a DNN is to estimate the similarity between its inner representations and those of another DNN.
This is challenging to quantify because networks are highly nonlinear and represent concepts in complex ways that may not reliably align with neurons or directions in activation space.
Nonetheless, a set of works have emerged to address this problem with a variety of both linear and nonlinear methods. 
These include single-neuron alignment \cite{li2015convergent, tatro2020optimizing, li2020representation, godfrey2022symmetries}, vector-space alignment \cite{wang2018towards}, canonical correlation analysis \cite{morcos2018insights}, singular vector canonical correlation analysis \cite{raghu2017svcca}, centered kernel alignment \cite{kornblith2019similarity, tang2020similarity, nguyen2020wide, raghu2021vision}, deconfounded representation similarity \cite{cui2022deconfounded}, layer reconstruction \cite{liang2019knowledge}, model stitching \cite{mcneely2020exploring, csiszarik2021similarity, bansal2021revisiting}, representational similarity analysis \cite{mehrer2020individual}, representation topology divergence \cite{barannikov2021representation}, and probing \cite{feng2020transferred}. 
Methods like these may aid in a better basic understanding of what features networks learn and how.
However, different methods often disagree about the extent to which layers are similar.
\cite{ding2021grounding} argue that these methods should be sensitive to changes that affect functional behavior and invariant to ones that do not. 
They introduce a benchmark for evaluating similarity measures and show that two of the most common methods, canonical correlation analysis and centered kernel alignment, each fail in one of these respects.

\section{Discussion}
\label{sec:discussion}

\noindent \textbf{Interpretability is closely linked with adversarial robustness research.}
There are several connections between the two areas, including some results from non-inner interpretability research.
(1) More interpretable DNNs are more robust to adversaries \cite{jyoti2022robustness}.
A number of works have studied this connection by regularizing the input gradients of networks to improve robustness \cite{ross2018improving, finlay2019scaleable, etmann2019connection, kim2019bridging, kaur2019perceptually, boopathy2020proper, hartl2020explainability, mangla2020saliency, du2021fighting, sarkar2021get, noack2021empirical}.
Aside from this, \cite{eigen2021topkconv} use lateral inhibition, and \cite{tsiligkaridis2020second} use a second-order optimization technique, each to improve \emph{both} interpretability and robustness. 
Furthermore, emulating properties of the human visual system in a convolutional neural network improves robustness \cite{dapello2020simulating}.
(2) More robust networks are more interpretable \cite{engstrom2019adversarial, augustin2020adversarial, ortiz2021optimism, elhage2022toy}.
Adversarially trained networks also produce better representations for transfer learning \cite{salman2020adversarially, agarwala2021one}, image generation \cite{santurkar2019image, casper2021robust, casper2022diagnostics}, modeling the human visual system \cite{engstrom2019learning}, and fitting symbolic graphs \cite{ren2021towards}.
(3) Interpretability tools can be used to design adversaries. Doing so is a way to rigorously demonstrate the usefulness of the interpretability tool.
This has been done by \cite{carter2019exploring, mu2020compositional, hernandez2021natural, casper2021robust, casper2022diagnostics} and has been used to more effectively generate adversarial training data \cite{ziegler2022adversarial}. 
As a means of debugging models, \cite{hubinger2019relaxed} argues for using ``relaxed'' adversarial training, which can rely on interpretability techniques to discover distributions of inputs or latents which may cause a model to fail. 
(4) Adversarial examples can be interpretability tools \cite{dong2017towards, tomsett2018failure, ilyas2019adversarial, casper2021robust, casper2022diagnostics} including adversarial trojan detection methods \cite{wang2019neural, guo2019tabor, gao2021design, liu2020survey, zheng2021topological, wang2022survey}. 

\medskip

\noindent \textbf{Interpretability is also closely linked with continual learning, modularity, network compression, and semblance to the human visual system.}
Continual learning methods involving parameter isolation, and/or regularization make neurons and weights more intrinsically interpretable. 
Sections \ref{subsec:continual_learning_weights} and \ref{subsec:continual_learning_neurons} discussed how these methods suggest intrinsic interpretations for weights and/or neurons.
Thus, they allow for each weight or neuron to be understood as having partial memberships in a set of task-defined \emph{modules}. 
Aside from this, a number of other intrinsic modularity techniques were the focus of Section \ref{subsec:modularity}.
And as discussed in Section \ref{subsec:modular_partitionings}, networks can also be interpreted by partitioning them into modules and studying each separately. 
Moreover, ``frivolous'' neurons, as discussed in Section \ref{subsec:frivolous neurons}, can include sets of redundant neurons that can be interpreted as modules. 
Networks with frivolous neurons are compressible, and compression can guide interpretations, and interpretations can guide compression, as discussed in Section \ref{subsec:frivolous neurons}.  
Finally, structuring networks to be more similar to the human visual system, including having convolutional filters that represent easily-describable patterns also improves robustness \cite{dapello2020simulating}. 

\medskip

\noindent \textbf{Interpretability techniques should scale to large models.}
Small networks and simple tasks such as MNIST classification \cite{lecun-mnisthandwrittendigit-2010} are often used for testing methods. 
However, simple networks performing simple tasks can only be deployed in a limited number of real-world settings, and they are sometimes easy to replace with other intrinsically interpretable, non-network models. 
As a result, the scalability of a technique is strongly related to its usefulness.
For example, capsule networks \cite{sabour2017dynamic} achieve impressive performance on MNIST classification and have intrinsic interpretability properties that convolutional networks lack.
However, they are much less parameter efficient and have thus far not achieved competitive performance beyond the CIFAR-10 \cite{krizhevsky2009learning} level, let alone the ImageNet \cite{russakovsky2015imagenet} level \cite{patrick2022capsule}.
Methods like these may offer excellent inspiration for future work, but if they fail to be tractable for large models, they will be of limited direct value for practical interpretability. 
We urge researchers to detail computational requirements and test the scalability of their methods.

\medskip

\noindent \textbf{Interpretability techniques generate hypotheses -- not conclusions. Producing merely-plausible explanations is insufficient. Evaluating validity and uncertainty are key.}
Mistaking hypotheses for conclusions is a pervasive problem in the interpretability literature \cite{lipton2018mythos, rudin2019stop, miller2019explanation}.
Consider the goal of explaining a particular neuron. 
There exist several methods to do so (Section \ref{sec:individual_neurons}). 
However, if such an approach suggests that the neuron has a particular role, this does not offer any guarantee that this explanation is complete and faithful to its true function.
Often, very plausible-seeming explanations do not pass simple sanity checks \cite{adebayo2018sanity} or are very easy to find counterexamples for \cite{olah2020zoom, bolukbasi2021illusion, hoffmann2021looks}.
A great number of works in interpretability have failed to go beyond simply inspecting the results of a method. 
More care is needed. 
Interpretability techniques can only be evaluated to the extent that they help users make testable predictions. 
They can only genuinely be useful inasmuch as those predictions validate.
And the validity of an interpretation is only granted on the distribution of data for which validating tests were conducted -- extrapolating interpretations is risky (e.g., \cite{bolukbasi2021illusion}). 
Developing specific methods for evaluating interpretability techniques is discussed later in Section \ref{sec:future_work}.

In addition to validity, it is important to quantify uncertainty.
Ideally, interpretations should be paired with confidence estimates.
How to measure certainty depends on the method at hand, but some approaches have been used such as supervising explanations (e.g., \cite{hendricks2016generating}), conducting multiple trials (e.g., \cite{olah2020zoom}), comparisons to random baselines (e.g., \cite{hod2021detecting, ravichander2020probing}), comparisons to other simple methods \cite{adebayo2018sanity}. or searching for cases in which an interpretation fails (e.g., \cite{bau2017dissection, bolukbasi2021illusion, hoffmann2021looks}).

\medskip

\noindent \textbf{Cherry-picking is harmful and pervasive. Evaluation of methods should not fixate on best-case performance.} 
Due to the inherent difficulty of interpreting DNNs, many works in the literature showcase individual, highly successful applications of their method, often in toy models. 
This can be useful for providing illustrative examples or specific insights.
But the evaluation of interpretability techniques should not be biased toward their best-case performance.
One hazard of doing this could be from overestimating the value of techniques.
And in fact, some works have found that certain methods only tend to perform well on a fraction of examples (e.g., \cite{bau2017dissection, locatello2019challenging, openai2019microscope, casper2019frivolous, voita2019analyzing, locatello2020sober, meister2021sparse, cammarata2020curve, hod2021detecting, bolukbasi2021illusion, meng2022locating, elhage2022solu}).

Another harm of cherrypicking might come from biasing progress in interpretability toward methods that fail to explain \emph{complex} subprocesses. 
Some methods are better equipped for this than others.
For example, attributing a feature's representation to a linear combination of neurons is strictly more general than attributing it to a single neuron.
It is likely that some kinds of features or computations in DNNs are more naturally human-understandable than others, so methods that are only useful for explaining simple subprocesses may be poorly-equipped for studying networks in general.

Works should evaluate how their techniques perform on randomly or adversarially sampled tasks.
For example, a work on characterizing neural circuits should not focus only on presenting results from circuits that were particularly amenable to interpretation. 
It should also aim to explain the role of randomly or adversarially sampled neurons inside of circuits or find circuits that can explain how the network computes randomly or adversarially selected subtasks.
If a method like this only succeeds in limited cases, this should be explicitly stated.

\medskip

\noindent \textbf{Ideally, progress in interpretability should neither decrease performance in general nor increase certain risky capabilities.}
On one hand, interpretable AI techniques should maintain competitiveness. 
It is key to avoid costs such as degraded task performance, increases in bias, higher compute demands, or difficulty to use in modern deep learning frameworks.
Competitive shortcomings like these could lead to ``value erosion'' \cite{dafoe2020ai} in which safer, more interpretable AI practices are not adopted in favor of more competitive approaches.

On the other hand, certain types of performance improvements from interpretability research may also be undesirable. 
Interpretability work should also not lead to increased capabilities if they make safety-related oversight more difficult.
For example, advances toward general intelligence may lead to serious harm if not managed appropriately \cite{bostrom2014superintelligence, muller2016future, tegmark2017life, russell2019human, christian2020alignment, ord2020precipice}.
One risky possibility is if interpretability is a byproduct of increased general capabilities. 
For example, large language models can often be prompted to ``explain'' their reasoning, but only as a result of having advanced, broad-domain abilities. 
Another way for this to occur is if interpretability leads to advancements in capabilities via basic model insights.
From the perspective of avoiding risks from advanced AI systems, neither of these is ideal. 
A focus on improving interpretability techniques \emph{without} commensurate increases in capabilities offers the best chance of preventing advancements in AI from outpacing our ability for effective oversight.
From this perspective, we argue that improvements in safety rather than capabilities should be the principal goal for future work in interpretability.

\section{Future Work} \label{sec:future_work}

\noindent \textbf{The connections between interpretability, modularity, adversarial robustness, continual learning, network compression, and semblance to the human visual system should be better understood.} 
One of the most striking findings of modern interpretability work is its connections with other paradigms in deep learning.
One of the central goals of this survey has been to highlight these connections (see Section \ref{sec:discussion}).
Currently, the intersections in the literature between interpretability and these other areas are relatively sparse.
Moving forward, an interdisciplinary understanding of interpretability may lead to insights and advancements spanning multiple domains.

\medskip

\noindent \textbf{Scaling requires efficient human oversight.} 
Many explanations obtained by state of the art interpretability techniques have involved a degree of human experimentation and creativity in the loop.
In some cases, many hours of meticulous effort from experts have been required to explain models or subnetworks performing very simple tasks (e.g., \cite{cammarata2020curve, nanda2022mechanistic}).
But if the goal is to obtain a thorough understanding of large systems, human involvement must be efficient.
Ideally, humans should be used for screening interpretations instead of generating them. 
Solutions can include using active learning (e.g., \cite{gao2020consistency}), weak supervision (e.g., \cite{boecking2020interactive}), implicit supervision using proxy models trained on human-labeled data (e.g., \cite{casper2021robust, casper2022diagnostics}), and/or rigorous statistical analysis of proxies (e.g., \cite{hod2021detecting, zhou2022exsum}) to reduce the need for human involvement.
Toward this end, obtaining additional high-quality datasets (e.g., \cite{bau2017dissection}) with richly-labeled samples may be valuable.

\medskip

\noindent \textbf{Focus on discovering novel behaviors -- not just analyzing them.} 
Many existing methods are only well-equipped to study how models behave in limited settings. 
For example, any interpretability method that relies on a dataset is limited to characterizing the model's behavior on that data distribution. 
But ideally, methods should not be limited to a given dataset or to studying potential failures when the failure modes are already known. 
For instance, an important practical problem is the detection of offensive or toxic speech, but no dataset contains examples of all types of offensive sentences, and having a human hand-specify a function to perfectly identify offensive from inoffensive speech is intractable.
Humans can, however, usually identify offensiveness when they see it with ease.

This highlights a need for techniques that allow a user to discover failures that may not be in a typical dataset or easy to think of in advance. 
This represents one of the \emph{unique} potential benefits of interpretability methods compared to other ways of evaluating models such as test performance. 
Toward this end, some inner interpretability methods that generate abstract understandings of subnetworks have proven to be useful (e.g., \cite{hernandez2021natural, mu2020compositional, santurkar2021editing, dai2021knowledge, meng2022locating}. 
However, methods based on \emph{synthesizing} adversarial examples may offer a particularly general approach for discovering novel failure modes (e.g., \cite{wang2019neural, guo2019tabor, casper2021robust, casper2022diagnostics}). 

\medskip

\noindent \textbf{Interpretability work may help better understand convergent learning of representations.}
Some works have hypothesized that similar features or concepts tend to occur across different model instances or architectures \cite{wentworth2022testing, olah2020zoom}.
Better understanding the extent to which systems learn similar concepts would lead to a more basic understanding of their representations and how interpretable we should expect them to be. 
If these hypotheses are true, interpreting one model in depth may be much more likely to lead to generalizable insights.
Continued work on measuring representational similarity between neural networks (see Section \ref{subsec:measuring_representational_similarity}) may be well-suited for making progress toward this goal. 



\medskip

\noindent \textbf{``Mechanistic interpretability'' and ``microscope AI'' are ambitious but potentially very valuable goals.} 
One direction for interpretability research is \emph{mechanistic interpretability}, which aims to gain an algorithmic-level understanding of a DNN's computations.
This can be operationalized as converting the DNN into some form of human-understandable pseudocode \cite{filan2018mechanistic}.
This is related to the goal of \emph{microscope} AI, which refers to gaining domain insights by thoroughly interpreting high-performing AI systems \cite{hubinger2020overview}. 
These capabilities would have advantages, including predicting counterfactual behaviours and reverse engineering models. 
Thus far, there have been a limited number of attempts towards this goal that have had some success by using small models, simple tasks, and meticulous effort from human experts \cite{verma2018programmatically, cammarata2020curve, elhage2022solu, nanda2022mechanistic}.
Future work in this direction may benefit by using techniques from program synthesis and analysis to automate the generation and validation of hypotheses.

\medskip

\noindent \textbf{Detecting deception and eliciting latent knowledge may be valuable for advanced systems.} 
A system is \emph{deceptive} if it passes false or incomplete information along some communication channel (e.g., to a human), despite having the capability to pass true and complete information.
Relatedly, \emph{latent knowledge} \cite{christiano2022eliciting} is something that a system ``knows'' but shows no signs of knowing. 
For example, a language model might babble a common misconception like ``bats are blind'' in some contexts despite having the ``knowledge'' that this is false.
Hidden knowledge like this may lead to deceptive behavior.
As an example, \cite{christiano2022eliciting} discusses a system that intentionally and deceptively manipulates the observations that a human sees for monitoring it.
In this case, knowledge about the true nature of the observations is latent.

Being able to characterize deceptive behavior and latent knowledge has clear implications for safer highly intelligent AI by allowing humans to know when a model may be untrustworthy.
But this may be difficult for several reasons including that (1) by definition, deceptive behavior and latent knowledge cannot be determined by observing the model's deployment behavior alone, (2) any mismatches between the features/concepts used by humans and the model require a method for ontology translation, and (3) it is unclear the extent to which a human can interpret an AI system that is superhuman on some task. 
However, inner interpretability methods offer a unique approach to these challenges via scrutinizing parts of the model's computational graph that may process latent knowledge.
Probing has shown potential for this \cite{burns2022discovering}.

\medskip

\noindent \textbf{Rigorous benchmarks are needed. Ideally, they should measure how helpful methods are for producing \emph{useful} insights that have relevance to engineers. These could involve rediscovering known flaws in networks.} 
Interpretability work on DNNs is done with numerous techniques, not all of which have the same end goal. 
For example, some methods aim to explain how a DNN handles a single input while others are aimed at a more generalizable understanding of it. 
For these reasons, plus the rapid development of techniques, widely-accepted benchmarks for interpretability do not yet exist.
This may be a limitation for further progress.
A well-known example of a benchmark's success at driving immense progress is how ImageNet \cite{russakovsky2015imagenet} invigorated work in supervised image classification. 

The weakest form of evaluation for an interpretability method is by its ability to merely suggest a particular characterization. 
For example, if feature synthesis is used to visualize a neuron, having a human look at the visualization and say ``this looks like X'' is a fraught basis for concluding that the neuron is an X-detector. 
This would be to conflate a hypothesis with a conclusion.
A somewhat more rigorous approach to evaluation is to make a simple testable prediction and validate it. 
For example, suppose the hypothesized X-detector activates more reliably for inputs that contain X than ones that do not. 
Another example is if the use of some method improves some quantitative proxy for interpretability (e.g., \cite{hod2021detecting}). 
This type of approach is valuable but still not ideal.

The end goal for interpretability tools should be to provide valid and useful insights, so methods for evaluating them ought to measure their ability to guide humans in doing \emph{useful} things with models.
In other words, interpretability tools should be useful for engineers, particularly ones who want to diagnose and debug models. 
Some works have made progress toward this. 
Examples include designing novel adversaries (e.g., \cite{geirhos2018imagenet, carter2019exploring, mu2020compositional, hernandez2021natural, ilyas2019adversarial, leclerc2021_3DB, casper2021robust, casper2022diagnostics, jain2022distilling, wiles2022discovering, ziegler2022adversarial, burns2022discovering}), manually editing a network to repurpose it or induce a predictable change in behavior (e.g., \cite{bau2018GAN, ghorbani2020shapley, wong2021leveraging, dai2021knowledge, meng2022locating}), or reverse-engineering a system using interpretability techniques (e.g., \cite{cammarata2020curve, elhage2021mathematical, nanda2022mechanistic}). 

One tractable approach for benchmarking suggested by \cite{hubinger_2021} would be to evaluate interpretability techniques by their ability to help a human find flaws that an adversary has implanted in a model.
Judging techniques by how well they help humans rediscover these flaws would offer a much more direct measure of their practical usefulness than ad hoc approaches to evaluation.
Related techniques for feature attribution methods have been argued for \cite{holmberg2022towards} and used \cite{adebayo2020debugging, bastings2021will, denain2022auditing} but have not yet popularized. 
Competitions for implanting and rediscovering flaws (e.g., \cite{mazeika2022trojan}) hosted at well-known venues or platforms may be a useful way to drive progress in both techniques and benchmarking. 

\medskip

\noindent \textbf{Combining techniques may lead to better results.} 
Interpretability techniques can often be combined.
For example, almost any intrinsic method could be used with almost any post hoc one.
However, the large majority of work in interpretability focuses on studying them individually.
Studying the interplay between methods is relatively unexplored. 
Some works have identified useful synergies (e.g., \cite{engstrom2019adversarial, wong2021leveraging})
But to the best of our knowledge, there are no works dedicated to thoroughly studying interactions between different methods. 
We hope that new baselines and increased demand for rigorously interpretable systems will further incentivize results-oriented interpretability work.

Consider an example. 
The ImageNet benchmark was very effective at advancing the state of the art on image classification performance in the 2010s. 
Over this time period, improvements in classification performance were not due to single techniques, but a combination of breakthroughs -- batch normalization, residual connections, inception modules, deeper architectures, etc. 
Similarly, we should not expect to best advance capabilities related to interpretability without a combination of methods. 

\medskip

\noindent \textbf{Applying interpretability techniques for debugging and debiasing in the wild.} 
Working to apply interpretability tools to find issues with real-world models (e.g. \cite{clark2022ai}) both helps for discovering issues in consequential applications and to test methods to see which ones may be the most practically useful.
In doing so, researchers should be critical of the ethical frameworks used in machine learning and particularly how they may diverge from the interests of people--particularly from disadvantaged groups--who may be the most adversely affected by these technologies \cite{birhane2022values}.

\medskip

\noindent \textbf{Growing the field of interpretability.} 
Many ethical or safety concerns with AI systems can be reduced via tools to better understand how models make decisions and how they may fail.
As a result, we argue that instead of being a separate interest, interpretability should be seen as a \emph{requirement} for systems that are deployed in important settings. 
As discussed above, a compelling path forward is via benchmarking and competitions.
There are some reasons for optimism.
The field is maturing, and a number of techniques have now proven their worth for practical insights and debugging.  
And although they are our focus here, we emphasize that that \emph{inner} interpretability methods will not be the only valuable ones for improving AI safety.

\medskip

\noindent \textbf{A paradigm shift toward engineering.} 
Currently, interpretability research produces few tools and insights that are useful in the real world. 
Being able to rigorously study the solutions learned by DNNs seems to have important potential for making DNNs safer, but they are currently rarely used for evaluation or engineering applications.
As discussed above, works in the literature too often treat hypotheses as conclusions \cite{lipton2018mythos, rudin2019stop, miller2019explanation} and fail to connect a method to useful applications. 
Some amount of exploratory work is clearly valuable for generating insights, and it should continue.
But the field has yet to produce many methods that are competitive in real applications. 
Motivations for interpretability work are ``diverse and discordant'' \cite{lipton2018mythos} and the term itself, as used in the literature, ``lacks precise meanings when applied to algorithms'' \cite{krishnan2020against}.
We join with \cite{doshivelez2017towards}, \cite{rudin2019stop}, \cite{miller2019explanation}, and \cite{krishnan2020against} in calling for grounding interpretability in meaningful applications. 
If interpretability tools are ultimately meant to help engineers diagnose and debug DNNs, the field should design and evaluate methods based on this.

We argue that moving forward, the most pressing change that is needed in the field is a focus on producing tools that are useful to engineers.
To better realize the potential of interpretability work for human-aligned AI, a more deliberate, interdisciplinary, and application-focused field will be important. 
It will be valuable to have more research that emphasizes diagnostics, debugging, adversaries, benchmarking, and leveraging useful combinations of different interpretability tools.

\section*{Acknowledgements} \label{acknowledgements}

We thank Davis Brown, Erik Jenner, Jan Kirchner, Max Lamparth, Ole Jorgensen, Florian Dorner, and Peter Hase for their feedback. Tilman Räuker is supported by the Long-Term Future Fund. 

\small
\bibliographystyle{plain}
\bibliography{bibliography.bib}

\end{document}